\ifdefined\pdfobjcompresslevel
\fi

\documentclass{jmlr}

\usepackage{longtable}

\usepackage{booktabs}
\usepackage[load-configurations=version-1]{siunitx} 

\usepackage{tikz}
\usetikzlibrary{positioning,arrows.meta,calc}

\theorembodyfont{\upshape}
\theoremheaderfont{\scshape}
\theorempostheader{:}
\theoremsep{\newline}

\jmlrvolume{}
\firstpageno{1}

\jmlryear{2026}
\jmlrproceedings{}{}
\jmlrworkshop{}

\title[WSBM-based ESN Comparison]{Analyzing the Dynamics of WSBM-based Echo State Networks using InputDSA}

 \author{\Name{Huy Mai} \Email{minhhuymaiduc@gmail.com}\\
 \addr Vietnam National University Ho Chi Minh City, University of Science
 \AND
 \Name{Robert Clarke} \Email{rc2205@bath.ac.uk}\\
 \addr University of Bath
 }

\begin{document}

\maketitle

\begin{abstract}
Bridging the structural gap between biological brains and artificial neural networks presents a pathway to altered computational dynamics. This paper introduces a Reservoir Computing framework that replaces random networks with biologically realistic modularity by applying the Weighted Stochastic Block Model (WSBM) to Echo State Networks (ESNs). This topological parameterization constructs reservoirs reflecting latent functional sub-circuits. Furthermore, the integration of Input Dynamical Similarity Analysis (InputDSA) provides a mechanism to isolate and compare how structural priors modulate intrinsic recurrent dynamics versus external inputs. By evaluating these architectures, we establish that structural diversity doesn't necessarily predicts behavioral diversity across the selected network subset. Furthermore, our demixed analysis confirms Anna Karenina principle of dynamics, demonstrating that environmental constraints enforce dynamical convergence in successful agents, while failing networks degenerate into idiosyncratic trajectories.
\end{abstract}


\section{Introduction}
\label{sec:intro}

Reservoir computing (RC), typically implemented as echo state networks (ESNs), provides an efficient approach to processing temporal data by leveraging the nonlinear dynamics of a fixed recurrent hidden layer. This architecture avoids the computational overhead of gradient-based optimization through time. However, standard ESNs rely on unstructured, random graphs (e.g., Erdős--Rényi) \citep{gauthierNextGenerationReservoir2021}, diverging from biological neural networks. Biological brains utilize organized meso-scale community structures to optimize information routing and processing efficiency \citep{betzelDiversityMesoscaleArchitecture2018}. By omitting these connectomic priors, artificial reservoirs fail to exploit the structural advantages of biological topologies.

To integrate biological plausibility into RC, we propose a connectome-constrained framework using the Weighted Stochastic Block Model (WSBM) \citep{aicherLearningLatentBlock2015, aicherAdaptingStochasticBlock2013}. The WSBM extracts latent block structures and weight distributions from real-world connectome data, enabling the construction of ESN reservoirs with meso-scale modularity. This prior replaces random initialization, imposing biological constraints on the dynamical system.

Evaluating these priors requires analyzing the network's internal temporal mechanisms rather than static spatial geometries. Because external inputs continuously perturb the reservoir's state, structural interpretation requires isolating this external forcing from intrinsic recurrence. We utilize Input Dynamical Similarity Analysis (InputDSA) \citep{huangInputDSADemixingThen2025}, which extends standard DSA \citep{ostrowGeometryComparingTemporal2023} via Subspace Dynamic Mode Decomposition with control (SubspaceDMDc). This method demixes intrinsic network dynamics from input-driven dynamics. By comparing the dynamical manifolds of random reservoirs against WSBM-ESNs, InputDSA provides a quantitative basis to demonstrate how connectome-derived topologies alter the temporal structure of computation.

The primary contributions of this paper are:
\begin{itemize}
    \item We introduce a generative framework that parameterizes ESN reservoirs using WSBM, embedding biological meso-scale modularity into the network topology.
    \item We apply InputDSA to isolate input-driven representations from intrinsic recurrent dynamics, providing qualitative evidence of how structural priors shape internal temporal flow
\end{itemize}

\section{Related works}
\label{sec:related}

A brief introduction to SBMs and ESNs is provided in Appendix \ref{apd:preliminary}.

\textbf{Reservoir Computing.} \quad 
Reservoir Computing (RC) emerged as a unifying paradigm \citep{Verstraeten2005ReservoirCW, VERSTRAETEN2007391} that consolidated models such as Echo State Networks (ESNs) and Liquid State Machines (LSMs) to address the training bottlenecks of standard Recurrent Neural Networks (RNNs). By decoupling the recurrent hidden layer—the reservoir—from the readout layer, temporal input signals are projected into a high-dimensional, non-linear state space through an untrained recurrent substrate. Because the reservoir remains fixed and only the linear readout requires training, RC is versatile; the recurrent substrate can be instantiated not only as artificial nodes in software, but also across diverse physical mediums, including photonic \citep{Vandoorne2008-ot}, electronic \citep{kulkarniMemristor}, and mechanical systems \citep{hauser2011towards}. While traditionally relying on uniform or Gaussian random graph models, recent work demonstrates that careful architectural design of the recurrent substrate yields quantifiable improvements. For instance, \cite{gallicchioReservoirTopologyDeep2019} showed that constrained reservoir topologies and structured organization of recurrent units can provide notable gains in predictive performance.

\textbf{Connectome-Constrained Networks.} \quad One way to constrain reservoir creation is through biologically grounded information. \cite{mcdanielInvestigatingEchoState2022} demonstrated that reservoirs endowed with biologically-inspired hierarchical structures achieve distinct performance characteristics compared to standard random graphs, and recent work by \cite{Milisav2026} further established that hierarchical modularity in neuromorphic reservoirs enhances memory capacity and supports multitasking. Efforts to integrate biological constraints into artificial networks generally fall into two categories: direct instantiation of empirical connectome weights \citep{MouseNet, Lappalainen2024}, and parameterized generative models that capture meso-scale topological rules \citep{betzel2017generative, Akarca2021, Salova2025-ni}. Our work aligns with the latter, seeking to incorporate the organizing principles of biological networks. 

\textbf{Weighted Stochastic Block Models.} \quad To capture these meso-scale biological rules, the Weighted Stochastic Block Model (WSBM) offers a robust generative framework. In network neuroscience, \cite{faskowitzWeightedStochasticBlock2018} demonstrated that WSBMs characterize the patterned interactions and community structures inherent in the human connectome across the lifespan. In the context of artificial computation, complex topological features resembling SBM structures have been shown to be computationally advantageous. \cite{dandreaComplexTopologicalFeatures2022} found that the modularity and link density of biological connectomes shape the learning performance of bio-inspired recurrent neural networks. Furthermore, \cite{dehghaniEvolutionaryOptimizationReveals2026} showed that when reservoir architectures are placed under evolutionary selection for spatiotemporal prediction, the evolved networks naturally conserve an SBM-like spectral envelope and lock macroscopic modularity to specific functional bands. Despite the established utility of WSBMs in describing brain networks and the recognized benefits of structural priors in ESNs, existing literature has not utilized the WSBM as a direct generative prior to jointly sample both the discrete topology and the continuous weight magnitudes of artificial reservoirs. That's the gap our paper aims to address.

\begin{figure}[tbp]
\floatconts
  {fig:wsbm-esn-overview}
  {\caption{Overview of the WSBM-based Echo State Network framework, with 4 main block models.}}
  {\includegraphics[width=\linewidth]{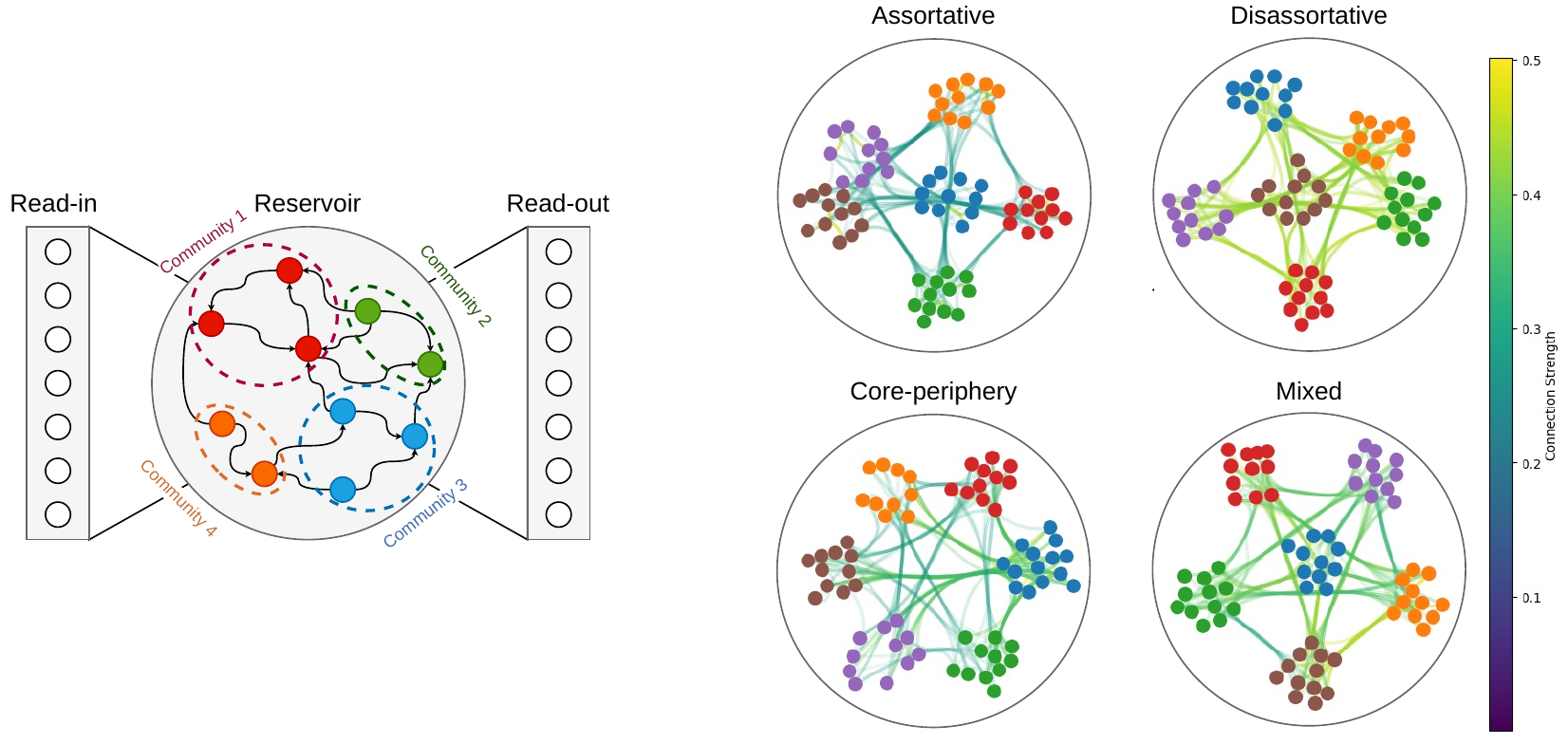}}
\end{figure}

\textbf{Representational and Dynamical System Analysis.} \quad While downstream performance metrics validate the utility of WSBM-generated reservoirs, they cannot justify how joint topological and weight priors alter internal information processing. To understand the computational mechanics of recurrent networks, it is necessary to examine their state-space trajectories. Historically, the geometric family of metrics—such as Representational Similarity Analysis (RSA) \citep{Kriegeskorte2008-wf}, Centered Kernel Alignment (CKA) \citep{CKA_og, kornblith2019similarityneuralnetworkrepresentations}, and Canonical Correlation Analysis (CCA) \citep{da6385d2-9c65-3860-bbcd-b821fdff69ff}—have been used to determine whether different neural systems represent stimuli with similar spatial arrangements in their latent spaces. However, because these methods operate primarily on static or time-averaged activity, they cannot distinguish between systems that possess identical representational geometry but execute different temporal computations. To address this limitation, dynamics-focused frameworks such as Dynamical Similarity Analysis (DSA) \citep{ostrowGeometryComparingTemporal2023} evaluate networks by embedding their temporal trajectories into linear dynamical operators (via Koopman operator theory) and aligning these dynamics matrices rather than static state-space geometries. Yet, standard DSA does not account for how input drive affects these trajectories, meaning it can conflate intrinsic network behavior with external forcing. Because our WSBM-ESNs are deployed in Reinforcement Learning environments—functioning as non-autonomous systems driven by environmental observations—we employ the recently proposed InputDSA framework \citep{huangInputDSADemixingThen2025} instead.

\section{Weighted Stochastic Block Model-based Echo State Network}
\label{sec:method}

\textbf{Generative model.} \quad We instantiate the exponential family of Section \ref{sec:sbm} as a normal distribution, so that the natural parameters $\eta_{z_i,z_j}$ correspond to a block mean and variance, and we retain the Bernoulli edge-existence term $\Theta$ of the binary SBM. A reservoir is then sampled from three symmetric $K \times K$ block matrices: the mean edge weight $\mu$, the weight dispersion $\sigma$, and the edge probability $\Theta$,
\begin{equation}
W_{ij} \sim \mathcal{N}\!\left(\mu_{z_i, z_j},\, \sigma^{2}_{z_i, z_j}\right) \cdot \text{Bernoulli}(\theta_{z_i, z_j}), \qquad W_{ii} = 0,
\label{eq:wsbm}
\end{equation}
with each undirected edge drawn once and mirrored so that $W = W^{\top}$, matching the undirected connectomes from which the model is estimated \citep{betzelDiversityMesoscaleArchitecture2018}. The sampled matrix is rescaled to a fixed spectral radius to enforce the echo state property. This joint parameterization is the point of departure from prior work: sampling topology and weight magnitude from a single generative prior, rather than imposing empirical weights directly \citep{MouseNet, Lappalainen2024} or generating topology alone \citep{betzel2017generative, Akarca2021}.

\textbf{Community interaction motifs.} \quad Following \cite{betzelDiversityMesoscaleArchitecture2018}, a pair of communities $\{r,s\}$ is classified by the ordering of three block means alone: it is \emph{assortative} if $\min(\mu_{rr}, \mu_{ss}) > \mu_{rs}$, \emph{disassortative} if $\mu_{rs} > \max(\mu_{rr}, \mu_{ss})$, and \emph{core--periphery} otherwise. Because $\mu$, $\sigma$ and $\Theta$ are independent, the model expresses architectures in which topology and weight structure carry \emph{different} motifs. Two consequences follow. First, a scalar edge probability renders every architecture topologically identical, confining all meso-scale structure to the weights; this degenerate case is what a single connection probability $p_{\mathrm{ER}}$ recovers. Second, the classification above reads means only and is therefore blind to $\sigma$, so dispersion constitutes a structural axis that the published taxonomy cannot resolve.

\textbf{Design space.} \quad The weight matrices admit one gauge freedom, since scaling $\mu$ and $\sigma$ by a common factor leaves the reservoir invariant after spectral rescaling. The design is accordingly parameterized by five structural axes: the signed weight contrast $c = (\mu_{\mathrm{in}} - \mu_{\mathrm{out}})/2$ expressed in units of $\bar{\sigma}$, the weight centre $m = (\mu_{\mathrm{in}} + \mu_{\mathrm{out}})/2$, the dispersion ratio $\sigma_{\mathrm{in}}/\sigma_{\mathrm{out}}$, the topology ratio $\varrho = \theta_{\mathrm{in}}/\theta_{\mathrm{out}}$, and the heterogeneity of community sizes. A signed $c$ subsumes the assortative and disassortative labels as opposite half-axes, with $c = 0$ recovering the Erdős--Rényi null. When sweeping $\varrho$, block probabilities are solved so that the expected edge count is held fixed, ensuring a topological motif is never confounded with edge abundance.

\section{Experiment}
\label{sec:experiment}

\textbf{Selection of a representative subset.} \quad Reservoirs are selected on \emph{structural} descriptors rather than on dynamical similarity, so that selection and outcome remain statistically independent; choosing systems for dynamical dissimilarity and subsequently reporting dynamical dissimilarity would be circular. Each sampled reservoir is characterized by eight descriptors computed from $W$: weighted modularity of the planted partition, mean participation coefficient, weighted clustering coefficient, global efficiency, the coefficients of variation of node strength and of betweenness centrality, the spectral ratio $|\lambda_2/\lambda_1|$, and the negative-weight fraction. Descriptors are standardized and projected by principal component analysis.

The population is stratified into five classes — assortative, disassortative, core--periphery, mixed, and the Erdős--Rényi null — and within each stratum $n = 20$ reservoirs are chosen by MaxMin selection in descriptor space, yielding $M = 100$. Stratification preserves the interpretability of the published taxonomy, while MaxMin selection within a stratum ensures the retained reservoirs span the within-class variation rather than clustering at its mode. The null stratum is a fifth arm of the design rather than a control appended after the fact, and it carries the burden of the central claim.

\textbf{Protocol.} \quad Each selected reservoir serves as the state encoder of a continuous control agent in a partially observable locomotion environment, with a model-free policy trained on the reservoir activations. All non-structural quantities — $N_x$, $\rho$, $\alpha$, $\gamma$, readout regularization and policy hyperparameters — are held constant across every reservoir, so that connectivity is the sole manipulated variable. RL tasks and codes are adapted from \cite{ermrl}\footnote{\url{https://github.com/corentinlger/ER-MRL}} under MIT license and only the developmental reinforcement loop is run; no evolutionary optimization of reservoir hyperparameters is performed.

\begin{figure}[tbp]
\floatconts
  {fig:pipeline}
  {\caption{Analysis pipeline. The dashed barrier marks the independence of selection from outcome: reservoirs are chosen on structural descriptors alone, and dynamical similarity is only ever a measured quantity, never a selection criterion.}}
  {%
  \begin{tikzpicture}[
    font=\small,
    stage/.style={draw,rounded corners,minimum height=1.0cm,text width=2.35cm,
                  align=center,inner sep=3pt},
    ar/.style={-{Latex[length=2mm]},thick}
  ]
  \node[stage] (s1) {\textbf{1.} Sample\\[1pt] \scriptsize design space};
  \node[stage,right=0.55cm of s1] (s2) {\textbf{2.} Describe\\[1pt] \scriptsize 8 graph descriptors};
  \node[stage,right=0.55cm of s2] (s3) {\textbf{3.} Stratify, MaxMin\\[1pt] \scriptsize $5 \times 20 = 100$};
  \node[stage,right=0.55cm of s3] (s4) {\textbf{4.} Train\\[1pt] \scriptsize RL, fixed hyperparams};
  \node[stage,right=0.55cm of s4] (s5) {\textbf{5.} InputDSA\\[1pt] \scriptsize $A$, $B$ distances};
  \foreach \a/\b in {s1/s2,s2/s3,s3/s4,s4/s5} \draw[ar] (\a) -- (\b);
  \draw[dashed,thick] ($(s3.north east)+(0.275,0.55)$) -- ($(s3.south east)+(0.275,-0.75)$);
  \node[font=\scriptsize] at ($(s3.north east)+(0.275,0.75)$) {\textit{selection} $\;\mid\;$ \textit{outcome}};
  \node[font=\scriptsize,below=0.1cm of s1] {structure only};
  \node[font=\scriptsize,below=0.1cm of s5] {dynamics only};
  \end{tikzpicture}%
  }
\end{figure}
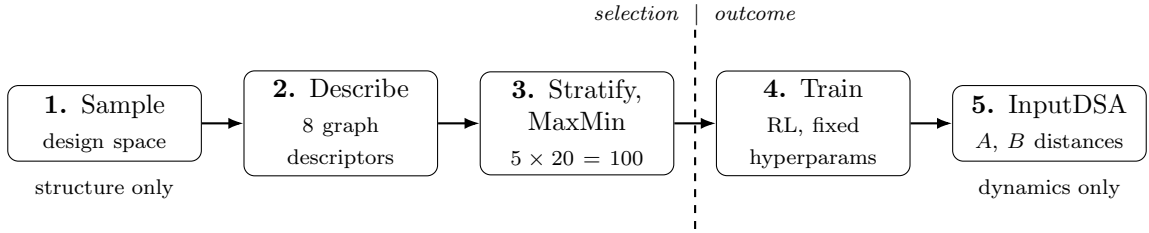

\textbf{Hypotheses.} \quad The study tests two claims:
\begin{itemize}
    \item[\textbf{H1}] \textbf{The Anna Karenina principle of dynamics.} Task constraints enforce dynamical convergence in top-performing agents, while failure modes promote idiosyncratic divergence. 
    \item[\textbf{H2}] \textbf{Structural diversity predicts behavioural diversity.} The descriptor-space spread of the selected subset predicts the spread of returns attained across it.
\end{itemize}

\section{Dynamics Analysis by InputDSA}
\label{sec:dsa}
\begin{figure}[htbp]
\floatconts
  {fig:top100-3-tasks}
  {\caption{Performance and dynamical representation of structured reservoir networks. \\ 
  A) Mean episode reward plot for 100 selected reservoirs compared to a Vanilla PPO baseline across three RL environments. \\ 
  B) Pairwise structural distance versus dynamical state distance for the top and bottom 10 reservoirs.}}
  {\includegraphics[width=\linewidth]{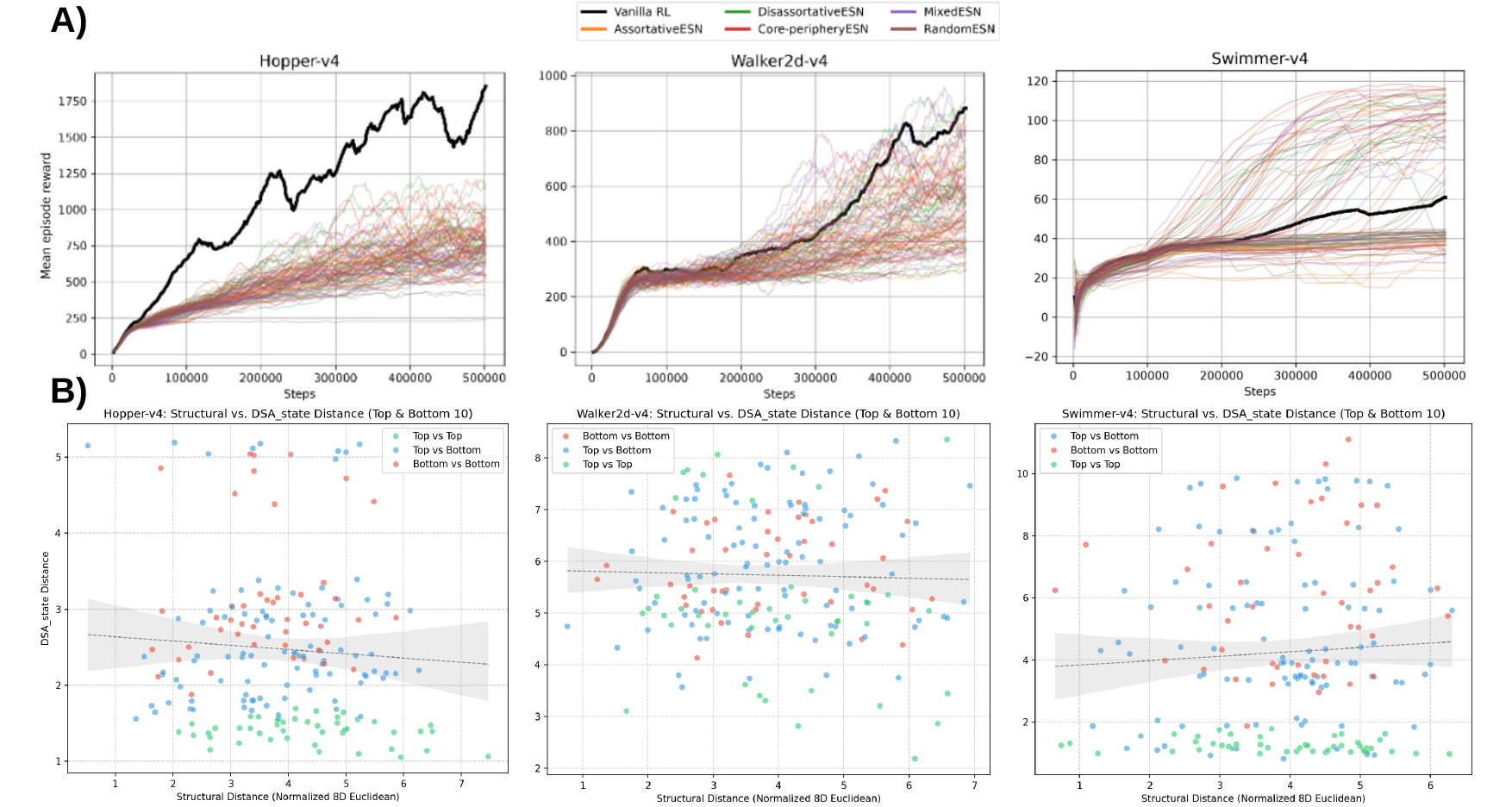}}
\end{figure}

We compare dynamics by fitting the non-autonomous state-space model $V(t+\delta t) = A V(t) + B s(t)$ using standard DMDc. Systems are then compared through the Procrustes distance between their controllability matrices, which decomposes into an intrinsic term, $\mathrm{DSA}_{\mathrm{state}}$, and an input-routing term, $\mathrm{DSA}_{\mathrm{input}}$ \citep{huangInputDSADemixingThen2025}. Here, $s(t)$ and $V(t)$ denote the system input and hidden state at time $t$, respectively. In our experiments, $s(t)$ concatenates the current environment observation $o_t$, previous action $a_{t-1}$, and previous reward $r_{t-1}$, while $V(t)$ is the context vector $c_t$ representing the reservoir state after processing $s(t)$, as illustrated in Fig.~\ref{fig:rl-loop}.

\textbf{Hypothesis 1}~

Our first hypothesis is built upon the Anna Karenina principle of dynamics: successful complex systems share common traits, while failures occur in unaligned ways. In reinforcement learning environments, achieving high rewards requires a specific set of computational responses. Consequently, top-performing networks are forced into a shared, optimal pattern of behavior to solve the task. Conversely, failing networks are unconstrained by the demands of success and diverge into scattered, uncoordinated trajectories. We hypothesize that successful models will exhibit aligned input-driven dynamics, whereas poor-performing models will display mismatched dynamics.

We present the performance of the 100 selected reservoirs in Fig.~\ref{fig:top100-3-tasks}A, which demonstrates significant fluctuations in mean episodic rewards for the Hopper-v4, Walker2d-v4, and Swimmer-v4 environments. This variability in reward is mirrored by a high variance in the corresponding network dynamics. However, when comparing this to the performance on the Ant-v4 and HalfCheetah-v4 tasks (Appendix~\ref{apd:second}, Fig.~\ref{fig:antv4-halfcheetahv4-fail}), the latter exhibit substantially more stable learning curves. This discrepancy implies that the specific environmental constraints and degrees of partial observability inherent to certain control tasks distinctly shape the variance of the resulting dynamics.  

To systematically assess this, we investigated the potential correlation between structural distance and the hidden state dynamics, quantified by $\mathrm{DSA}_{\mathrm{state}}$. As illustrated in Fig.~\ref{fig:top100-3-tasks}B, fitting a linear model reveals no significant correlation between the structural descriptors and the intrinsic dynamical state representations. However, a critical observation emerges from the input-driven dynamics: the structural variance among the most successful models is high, yet their dynamical dissimilarity is low. This suggests that the environments impose strict computational constraints that enforce dynamical convergence regardless of the underlying structural topology.  

This convergence aligns with recent findings utilizing InputDSA on control tasks like olfactory plume tracking (see Section 4.~Applications~\cite{huangInputDSADemixingThen2025}). In these settings, top-performing agents develop similar input-driven dynamics because success depends on coordinated responses to environmental stimuli. Furthermore, analyzing the singular value spectrum of the input-mapping operator reveals that inputs successfully excite a broader range of dimensions within the state space of top agents, whereas bottom-performing agents diverge toward heterogeneous dynamics. Thus, the shared demands of the task force high performance WSBM-ESNs to converge onto an optimal, input-responsive dynamical manifold.

\begin{figure}[tbp]
\floatconts
  {fig:top-bottom-10-swimmerv4}
  {\caption{InputDSA distance matrices comparing the top 10 (T\_) and bottom 10 (B\_) performing reservoir architectures for the Swimmer-v4 task. The pairwise dynamical distances are computed using five metric configurations: Joint, State (Joint), Control (Joint), State (Separate), and Control (Separate).}}
  {\includegraphics[width=\linewidth]{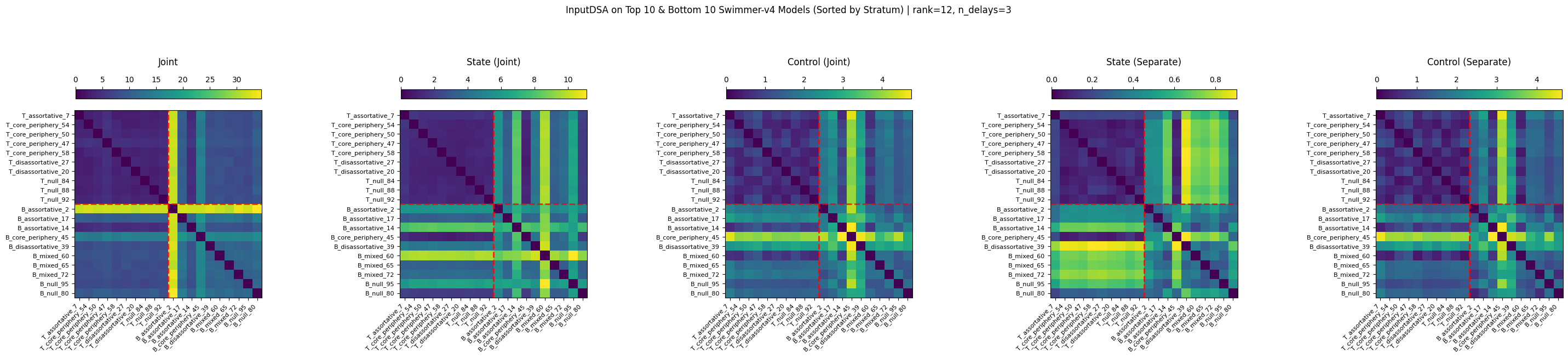}}
\end{figure}

While recent literature suggests SubspaceDMDc (via N4SID) is superior for partial observable systems due to its handling of unobserved state biases, we chose to utilize standard DMDc for our WSBM-ESN analysis. When applying N4SID to the high-dimensional, coupled internal states of our reservoirs (Fig~\ref{fig:n4sid-fail}), the algorithm exhibited numerical instability. Because ESNs operate with recurrent matrices scaled near the edge of chaos, the subspace projection steps seemingly produced near-singular matrices, causing the Joint metric distances to exceed $10^6$. In contrast, as featured from Fig~\ref{fig:top-bottom-10-swimmerv4}, standard DMDc provided stable, bounded alignments suitable for our observable reservoir states.

In general, we observed patterns of similarity across the five InputDSA matrices that validate our hypothesis. On the Joint metric, the low pairwise distances among the top models indicate a strong overall alignment of the non-autonomous systems. On the State (Joint and Separate) metrics: This tight alignment reveals that successful WSBM-constrained reservoirs converge onto an optimal intrinsic dynamical manifold to solve the task. On the Control (Joint and Separate) metrics: The high variance and scattered distances among the bottom models highlight that bottom agents suffer from chaotic, uncoordinated input-routing, failing to integrate environmental observations. In conclusion, these metrics confirm the Anna Karenina principle of dynamics hypothesis.

\textbf{Hypothesis 2}~

Our second hypothesis concerns whether the structural diversity engineered into the selected subset propagates into behavioural diversity. MaxMin selection maximizes descriptor-space spread within each stratum, so if the eight descriptors capture the axes of connectivity that govern computation, reservoirs lying far apart in descriptor space should attain measurably different returns. We therefore test the pairwise form of the claim: whether the descriptor distance between two reservoirs predicts the difference in the returns they achieve.
Behavioural diversity is substantial. Evaluating all M = 100 reservoirs with deterministic rollouts, returns span 221–3010 on Hopper-v4, 124–1891 on Walker2d-v4, and 17.8-128.7 on Swimmer-v4, with a coefficient of variation strikingly stable across the three environments (CV = 0.54, 0.50, 0.52). The selected reservoirs are therefore far from behaviourally interchangeable, and the environments differ in the scale of return they support rather than in the relative spread they induce.

That diversity is not, however, organized by position in descriptor space. Across all 4950 reservoir pairs, the rank correlation between descriptor distance and absolute return difference is indistinguishable from zero in every environment (Spearman $|\rho| \le 0.012$, all $p \ge 0.38$). Restricting to within-stratum pairs, which removes between-class variation, yields no coherent pattern either: two of fifteen tests reach nominal significance, in inconsistent directions across environments. We read this as a statement about the descriptor basis rather than about structure itself. Connectivity evidently matters for the returns attained, since the reservoirs differ by more than an order of magnitude, but the eight descriptors of Section 4 do not span the axis along which it matters. A further limitation is that each reservoir contributes a single training run, so the observed spread confounds the structural contribution with the stochasticity of policy optimization; separating the two requires replicate seeds per reservoir, which we identify as the primary extension of this analysis.
\section{Conclusion and Discussion}
\label{sec:conclusion}
This study presents a generative framework that parameterizes Echo State Networks (ESNs) using the Weighted Stochastic Block Model (WSBM) to embed biologically realistic meso-scale modularity. By applying Input Dynamical Similarity Analysis (InputDSA), we established that environmental task constraints enforce dynamical convergence in successful networks, whereas failing models diverge into uncoordinated trajectories.  
We have relegated the plots of the Ant-v4 and HalfCheetah-v4 tasks to Appendix \ref{apd:second}. The mean episodic reward plots for these specific environments are stable and exhibit low variance. Consequently, the dynamics across all generated systems converge in a similar manner under these specific environmental constraints. Examining these stable examples provides a clear baseline, and the mechanisms driving this uniform convergence warrant further investigation.  

The choice of dynamical decomposition algorithm presents a methodological limitation. Although N4SID is theoretically suited for partially observable systems, it exhibited numerical instability when applied to the high-dimensional internal states of our reservoirs. As a result, standard DMDc was utilized to ensure stable, bounded alignments. Future research should investigate adaptations to N4SID or alternative subspace methods that can reliably handle the specific dynamics of ESNs

Finally, future work should refine the information routing mechanism of the read-in layer. Rather than employing standard random routing, directing external inputs to specific structural modules within the reservoir would allow for a more precise evaluation of how localized, meso-scale topologies process targeted stimuli.


\bibliography{pmlr-sample}

\appendix
\section{Preliminaries}\label{apd:preliminary}

\subsection{Echo State Network}
\label{sec:esn}

Echo State Networks (ESNs) \citep{Jaeger2001TheechoST} provide a recurrent neural network architecture where temporal dynamics are modeled using a fixed, nonlinear hidden layer, termed the reservoir. By restricting training to the output readout layer via linear regression, ESNs avoid the computational expense of backpropagation through time. 

Let $s(t) \in \mathbb{R}^{N_u}$ denote the input vector and $V(t) \in \mathbb{R}^{N_x}$ denote the reservoir state at discrete time $t$, where $N_x$ is the number of reservoir units. In a standard ESN, the reservoir topology is defined by a sparse random graph, often instantiated as an Erdős--Rényi network with connection probability $p_{\mathrm{ER}}$. The state evolution of the reservoir is governed by the dynamical equation:
\begin{equation*}
V(t+\delta t)=(1-\alpha)V(t)+\alpha f\!\left(\gamma W_{\mathrm{in}}s(t)+\rho W V(t)\right),
\end{equation*}
where $\alpha = \delta t/\tau \in (0, 1]$ represents the leakage rate governed by the integration timescale $\tau$. The function $f(\cdot)$ is an element-wise nonlinear activation (e.g tanh function), $W_{\mathrm{in}} \in \mathbb{R}^{N_x \times N_u}$ is the input weight matrix, and $W \in \mathbb{R}^{N_x \times N_x}$ is the fixed recurrent weight matrix defining the network's internal connectivity. The scalar multipliers $\gamma$ and $\rho$ dictate the input scaling and recurrent scaling, respectively.

For stable functioning, the dynamical system must satisfy the echo state property, which asserts that the asymptotic state of the reservoir depends on the history of the driving input $s(t)$, independent of initial conditions: $\lim_{t \to \infty} \|V(t, s, V_0) - V(t, s, V_0')\| = 0$. Hence, it is necessary that the recurrent weight matrix $W$ got its spectral radius scaled, $\max_{i} |\lambda_i(\rho W)| < 1$, with $\lambda_i(\rho W)$ denote the $i$-th eigenvalue of the scaled recurrent weight matrix in the complex plane. \citep{JAEGER2007335, lukoseviciusPracticalGuideApplying2012}.

\subsection{Stochastic Block Model}
\label{sec:sbm}

The Stochastic Block Model (SBM) \citep{HOLLAND1983109, Bridgeford_Loftus_Vogelstein_2025} is a generative probabilistic model formulated to infer latent community structures within a network. Let a network of $N$ nodes be represented by an adjacency matrix $A \in \mathbb{R}^{N \times N}$. The SBM partitions the node set into $K$ disjoint blocks, with the latent block assignment for node $i$ denoted by $z_i \in \{1, \dots, K\}$. The premise of the SBM is stochastic equivalence: the probability of an edge between any two nodes depends exclusively on their respective block assignments. Formally, the existence of a binary edge $A_{ij} \in \{0, 1\}$ is modeled as a conditionally independent Bernoulli random variable:
\begin{equation*}
A_{ij} \mid z_i, z_j \sim \text{Bernoulli}(\theta_{z_i, z_j}),
\end{equation*}
where $\Theta \in [0, 1]^{K \times K}$ is the inter-block connection probability matrix, and $\theta_{z_i, z_j}$ defines the probability of an edge forming between a node in block $z_i$ and a node in block $z_j$.
A limitation of the standard SBM is its restriction to binary topologies, which excludes the parameterization of interaction strengths inherent to empirical connectomes. The Weighted Stochastic Block Model (WSBM) \citep{aicherLearningLatentBlock2015, aicherAdaptingStochasticBlock2013} resolves this by extending the framework to jointly model the topological structure and continuous edge weights. Under the WSBM formulation, the conditional distribution of an edge weight is parameterized via a univariate exponential family:
\begin{equation*}
A_{ij} \mid z_i, z_j \sim \text{ExponentialFamily}(\eta_{z_i, z_j}),
\end{equation*}
where $\boldsymbol{\eta} \in \mathbb{R}^{K \times K}$ represents the matrix of natural parameters, such that its element $\eta_{z_i, z_j}$ governs the weight distribution between communities $z_i$ and $z_j$. This is expressed via the probability density function $P(A_{ij} \mid z_i, z_j) = h(A_{ij}) \exp(\eta_{z_i, z_j}^T T(A_{ij}) - c(\eta_{z_i, z_j}))$, where $T(\cdot)$ is the sufficient statistic and $h(\cdot)$ is the base measure. This extension provides the formal apparatus required to map the continuous, meso-scale organization of biological connectomes into the topological priors $W$ of the aforementioned Echo State Networks.

\clearpage
\section{Hyperparameter space and failure mode}\label{apd:second}

\begin{figure}[htbp]
\floatconts
  {fig:hyperparam-space}
  {\caption{Hyperparameters space selection.}}
  {\includegraphics[width=\linewidth]{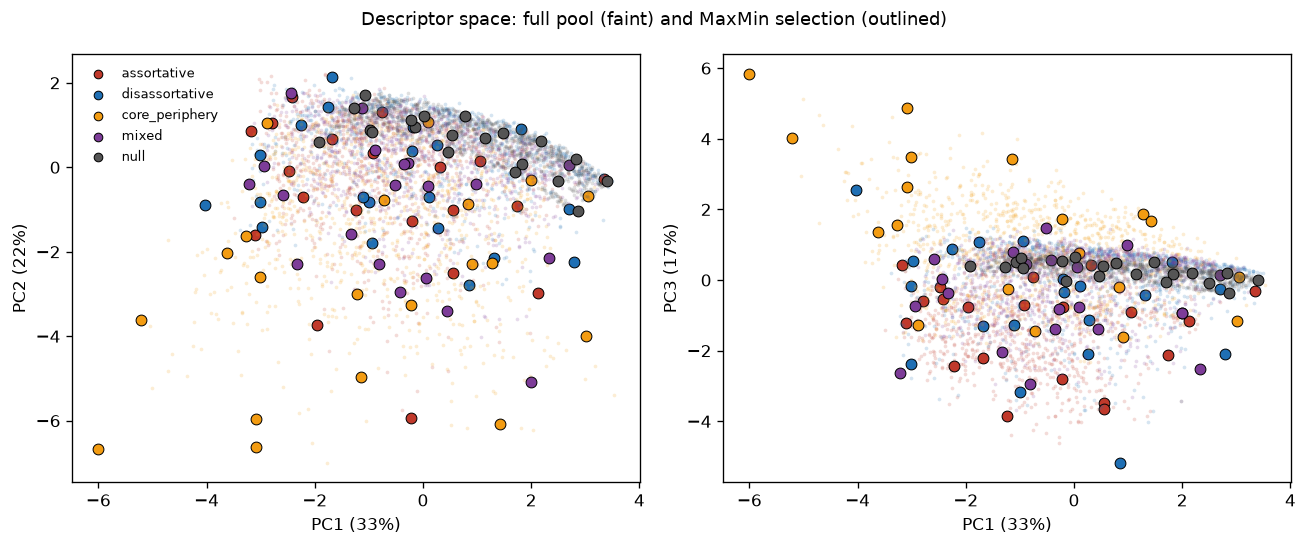}}
\end{figure}

\begin{figure}[htbp]
\floatconts
  {fig:fail-mackey-glass-5-sytems}
  {\caption{Failure mode of InputDSA on each instance of 5 classes on Mackey-Glass timeseries due to input's low-dimensionality vs OU noise timeseries.}}
  {\includegraphics[width=\linewidth]{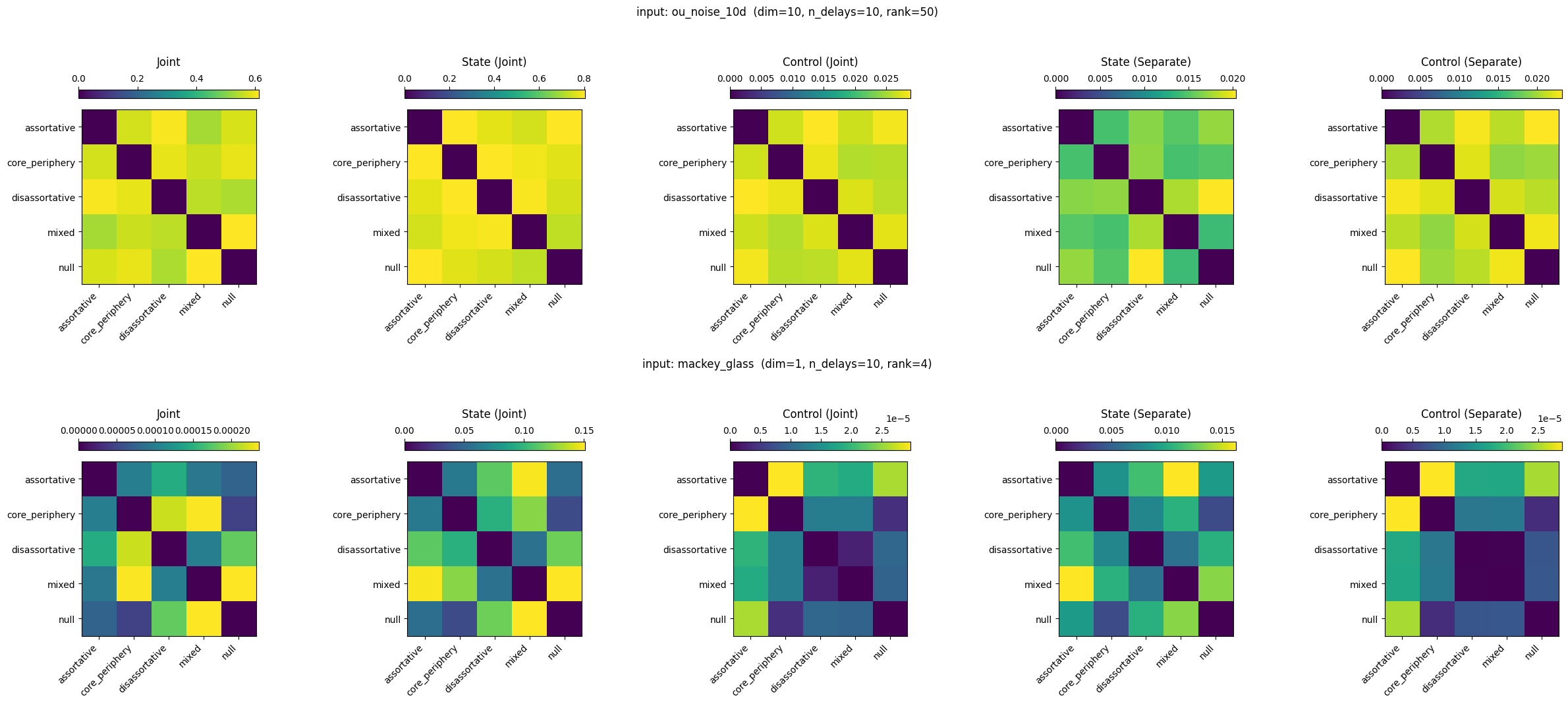}}
\end{figure}

\begin{figure}[htbp]
\floatconts
  {fig:antv4-halfcheetahv4-fail}
  {\caption{The performance of agents across Ant-v4 and HalfCheetah-v4 tasks with its corresponding top and bottom 10 dynamics}}
  {\includegraphics[width=\linewidth]{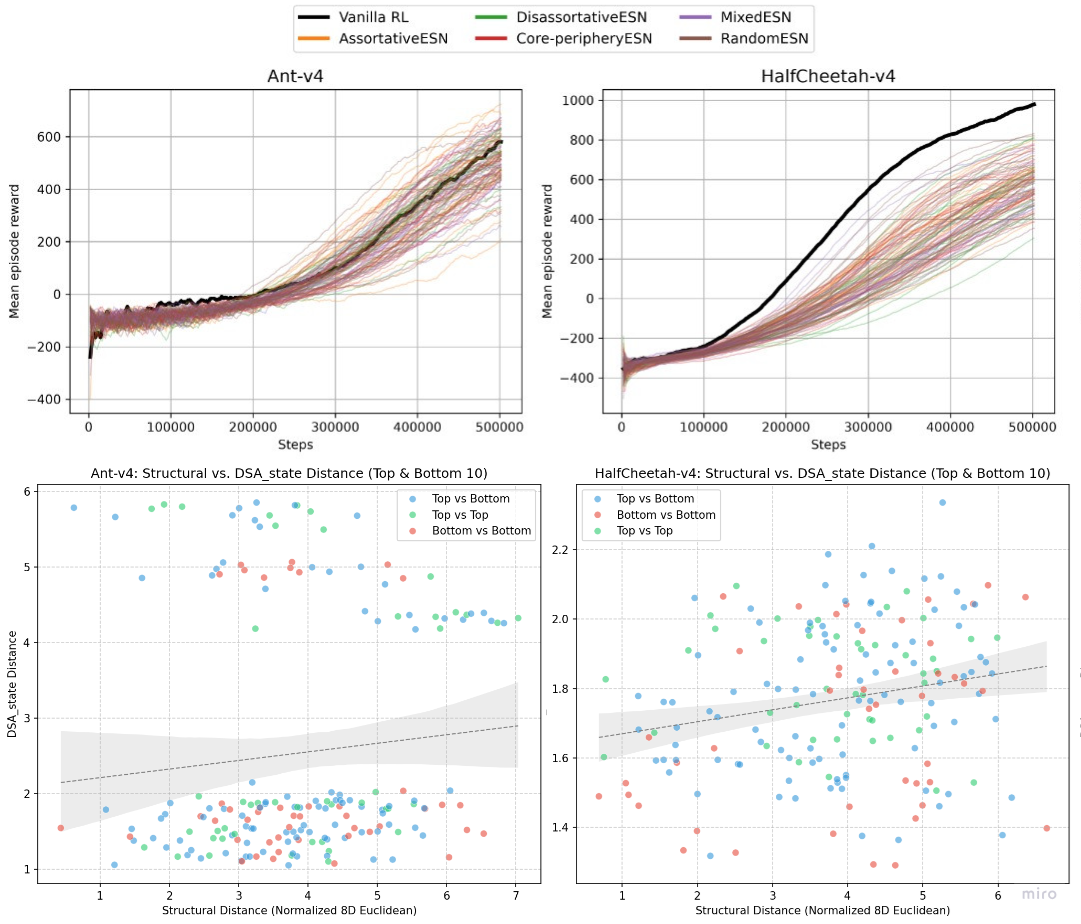}}
\end{figure}

\begin{figure}[htbp]
\floatconts
  {fig:n4sid-fail}
  {\caption{Failure mode of using N4SID on Swimmer InputDSA.}}
  {\includegraphics[width=\linewidth]{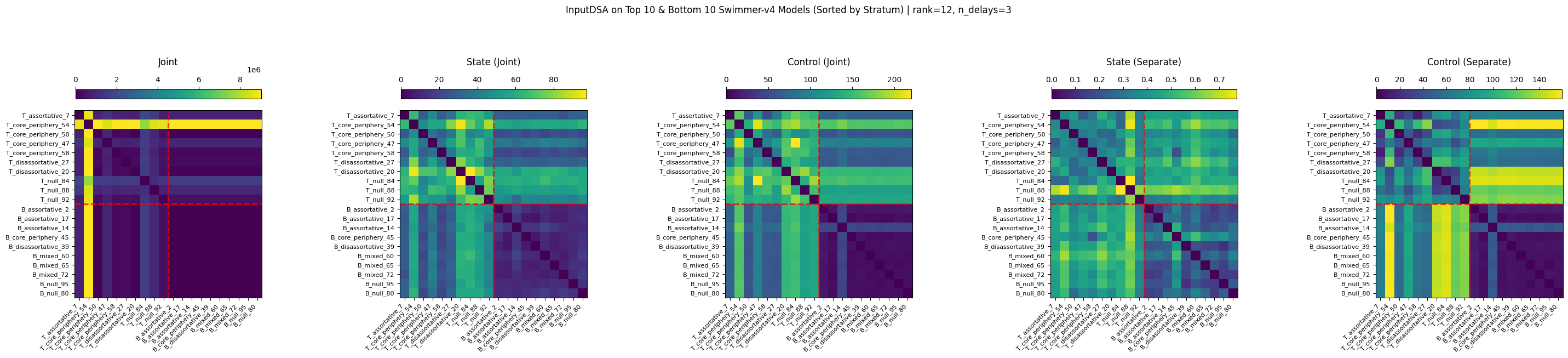}}
\end{figure}


\clearpage
\section{}
\subsection{Algorithm}
\begin{algorithm}\label{algo:init}
\vspace{2mm}
\hrule
\vspace{2mm}
\textbf{Input:} Number of nodes $N$, communities $K$, target spectral radius $\rho$, negative edge probability $p_{\text{neg}}$, community assignment logic. Block matrices for means $\Omega \in \mathbb{R}^{K \times K}$, standard deviations $\Sigma \in \mathbb{R}_{+}^{K \times K}$, and connection probabilities $\Theta \in [0,1]^{K \times K}$.\\
\textbf{Output:} Initialized reservoir weight matrix $W_{\text{res}} \in \mathbb{R}^{N \times N}$, community assignment vector $z \in \{1, \dots, K\}^N$.
\vspace{2mm}
\hrule
\vspace{2mm}

\begin{enumerate}
    \item \textit{Determine community assignments}
    \begin{enumerate}
        \item[(a)] $z \leftarrow$ Equal distribution, Dirichlet distribution, or explicit array configuration mapping each node $i$ to a community $k$.
    \end{enumerate}
    
    \item \textit{Sample initial states and connectivity masks}
    \begin{enumerate}
        \item[(a)] $W_{ij} \sim \mathcal{N}(\Omega_{z_i, z_j}, \Sigma_{z_i, z_j}^{2})$ for all $i, j \in \{1, \dots, N\}$
        \item[(b)] $M_{ij} \sim \text{Bernoulli}(\Theta_{z_i, z_j})$ for all $i, j \in \{1, \dots, N\}$
    \end{enumerate}
    
    \item \textit{Enforce undirected topology and self-connection constraints}
    \begin{enumerate}
        \item[(a)] $W_{ij} \leftarrow W_{ij} \cdot M_{ij}$ for $i < j$
        \item[(b)] $W_{ji} \leftarrow W_{ij}$ for $i < j$
        \item[(c)] $W_{ii} \leftarrow 0$ for all $i \in \{1, \dots, N\}$
    \end{enumerate}
    
    \item \textit{Apply structural perturbations}
    \begin{enumerate}
        \item[(a)] $F_{ij} \sim \text{Bernoulli}(p_{\text{neg}})$ for $i < j$
        \item[(b)] $W_{ij} \leftarrow -|W_{ij}|$ and $W_{ji} \leftarrow -|W_{ij}|$ for all $\{i, j \mid i < j \land F_{ij} = 1\}$
    \end{enumerate}
    
    \item \textit{Scale to target spectral dynamics}
    \begin{enumerate}
        \item[(a)] $\lambda_{\text{max}} \leftarrow \max |\text{eigvals}(W)|$
        \item[(b)] $W_{\text{res}} \leftarrow \rho \frac{W}{\lambda_{\text{max}}}$
    \end{enumerate}
\end{enumerate}
\vspace{2mm}
\hrule
\end{algorithm}
\subsection{RL Loop}
\begin{figure}[hbp]
\floatconts
  {fig:rl-loop}
  {\caption{A diagram features the flow of information during a RL task, adapted from \cite{ermrl}}}
  {\includegraphics[width=\linewidth]{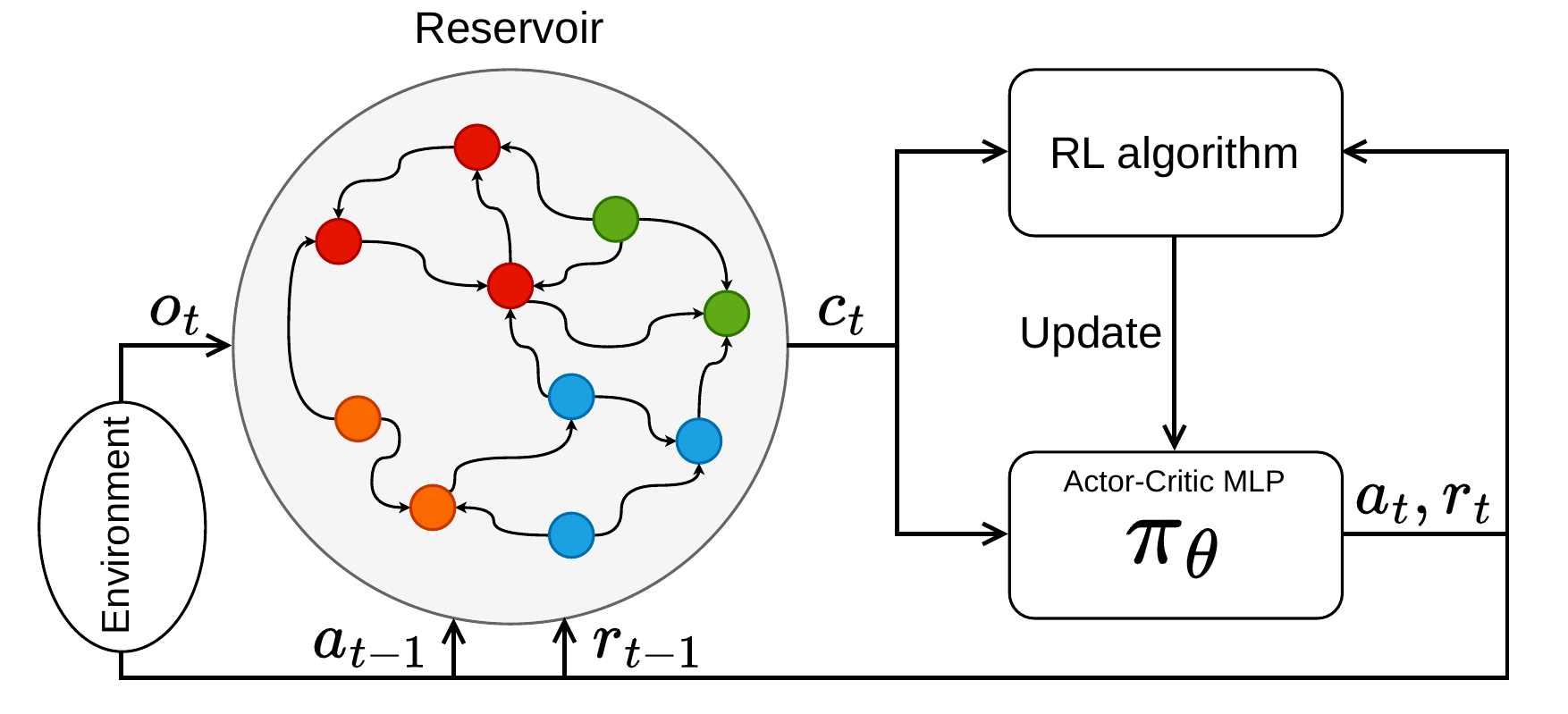}}
\end{figure}

\end{document}